\documentclass{article}

\usepackage{arxiv}

\usepackage[utf8]{inputenc} 
\usepackage[T1]{fontenc}    
\usepackage{hyperref}       
\usepackage{url}            
\usepackage{booktabs}       
\usepackage{amsfonts}       
\usepackage{nicefrac}       
\usepackage{microtype}      
\usepackage{lipsum}
\usepackage{graphicx}
\usepackage{amsmath}
\usepackage{caption}

\usepackage{wrapfig}

\usepackage{algorithm,algorithmicx,algpseudocode}

\graphicspath{ {./fig/} }

\title{Real-World Denoising via Diffusion Model}

\author{
  Cheng Yang\\
  School of Mathematical Sciences\\
  Dalian University of Technology\\
  Dalian, Liaoning 116024 \\
  \texttt{yangcheng1@mail.dlut.edu.cn} \\
    \And
  Lijing Liang \\
  School of Mathematical Sciences\\
  Dalian University of Technology\\
  Dalian, Liaoning 116024 \\
  \texttt{lljydx@mail.dlut.edu.cn } \\
   \And
  Zhixun Su\\
  School of Mathematical Sciences\\
  Dalian University of Technology\\
  Dalian, Liaoning 116024 \\
  \texttt{zxsu@dlut.edu.cn} \\
}

\begin{document}
\maketitle
\begin{abstract}
Real-world image denoising is an extremely important image processing problem, which aims to recover clean images from noisy images captured in natural environments. In recent years, diffusion models have achieved very promising results in the field of image generation, outperforming previous generation models. However, it has not been widely used in the field of image denoising because it is difficult to control the appropriate position of the added noise. Inspired by diffusion models, this paper proposes a novel general denoising diffusion model that can be used for real-world image denoising. We introduce a diffusion process with linear interpolation, and the intermediate noisy image is interpolated from the original clean image and the corresponding real-world noisy image, so that this diffusion model can handle the level of added noise. In particular, we also introduce two sampling algorithms for this diffusion model. The first one is a simple sampling procedure defined according to the diffusion process, and the second one targets the problem of the first one and makes a number of improvements. Our experimental results show that our proposed method with a simple CNNs Unet achieves comparable results compared to the Transformer architecture. Both quantitative and qualitative evaluations on real-world denoising benchmarks show that the proposed general diffusion model performs almost as well as against the state-of-the-art methods.
\end{abstract}

\section{Introduction}
Images are an important source of external information for humans and form the basis of human vision, containing a large amount of information about objects. However, during the process of image acquisition, transmission, and storage, images are often corrupted by unwanted signals, resulting in a degradation of image quality that can be detrimental to subsequent image processing operations. The purpose of real-world image denoising is to remove as much unnecessary or redundant noise as possible to ensure the integrity of the original information in the image and to facilitate the application of high-quality images to higher-level computer vision tasks. Real-world image denoising to obtain high quality images is the basis for correct recognition of image information and has a facilitating effect on the performance of other processing aspects of digital images. High-quality, clear images can help solve specific problems in medical imaging, autonomous driving, pattern recognition, and other areas. Therefore, real-world image denoising and image quality improvement are prerequisites for image recognition and various feature extraction.

With its powerful nonlinear adaptation capability, convolutional neural networks
(CNNs) perform well in many low-level computer vision tasks~\cite{Dong2014ImageSU, Schmidt2014ShrinkageFF, Kim2015AccurateIS, Nah2016DeepMC}. Traditional real-world image denoising algorithms use hand-crafted features and mathematical models to remove noise from an image. With CNNs, however, the network can learn these features on its own, without any prior knowledge of image processing techniques or statistical models. Convolutional neural networks can map real-world noisy images to noise-free images in image denoising methods, which is one of the classical algorithms for real-world image denoising, and the performance of convolutional neural networks~\cite{Jain2008NaturalID} is comparable to that of the most advanced denoising algorithms based on wavelet transform and Markov random field~\cite{Malfait1997WaveletbasedID}. There are two categories of image denoising methods based on convolutional neural networks: multi-layer perceptron models~\cite{Burger2012ImageDC}and deep learning methods. Autoencoder~\cite{Vincent2008ExtractingAC} is a classical image denoising method based on the multi-layer perceptron model. The denoising autoencoder mentioned above can be used for pre-training deep neural networks just like ordinary autoencoders, but the denoising autoencoder can add noise to the original input as the input of the encoder, which can force the deep neural network to learn more robust invariant features so that the input can be more effective. TNRD~\cite{Chen2015TrainableNR} is a feed-forward deep neural network, also called trainable nonlinear reaction-diffusion model, which has achieved better denoising performance. Convolutional neural networks are used to extract features, which can reduce the influence of noise while extracting features, and then complete image inpainting by deconvolution to achieve the purpose of image denoising\cite{Mao2016ImageRU}. FC-AIDE~\cite{Cha2018FullyCP} based on contextual pixel-level mapping, uses a fully convolutional enhanced supervised model to achieve more robust adaptability and improve image denoising performance through regularization methods. By exploiting the powerful capabilities of convolutional neural networks, these methods have achieved significant success in real-world image denoising.

GCBD~\cite{Chen2018ImageBD} uses the generative adversarial networks~\cite{Goodfellow2014GenerativeAN} for blind denoising, first a GAN network is trained to estimate the noise distribution on the input noisy image and generate noisy samples. Second, a paired training dataset is constructed using the noise patches collected in the first step, and then a deep CNN is trained to denoise the given noisy image. DnCNN~\cite{Zhang2016BeyondAG} uses residual learning in the field of image denoising for the first time, and combines residual learning with batch normalization to speed up the training of the image denoising model and improve the denoising effect. DnCNN can effectively remove uniform Gaussian noise and suppress noise within a certain noise level. However, the real-world noise is usually due to the characteristics of color channel correlation and signal dependency, which cannot be handled by uniform Gaussian noise. In this case, FFDNet~\cite{Zhang2017FFDNetTA} tackles complex noise in real scenes by using noise level estimation as the input to the network, thus preserving more details in the image. CBDNet~\cite{Guo2018TowardCB} proposes a framework that includes a noise estimation subnetwork and a non-blind denoising subnetwork, which combines the noise estimation and non-blind denoising models, and uses synthetic noise and real noise images for network training to improve denoising performance. Convolutional neural networks can automatically extract image features and reduce computational cost during the denoising process, while avoiding complicated computations during learning and inference. However, real-world image denoising methods based on convolutional neural networks have a remarkable effect on images with additive white Gaussian noise and struggle to deal well with real-world noise images. In addition, the convolution kernels of convolutional neural networks cannot adapt to the image content when long-term dependencies need to be established, resulting in global information loss.

\begin{figure*}
	\centering	         
	\includegraphics[width=0.8\textwidth]{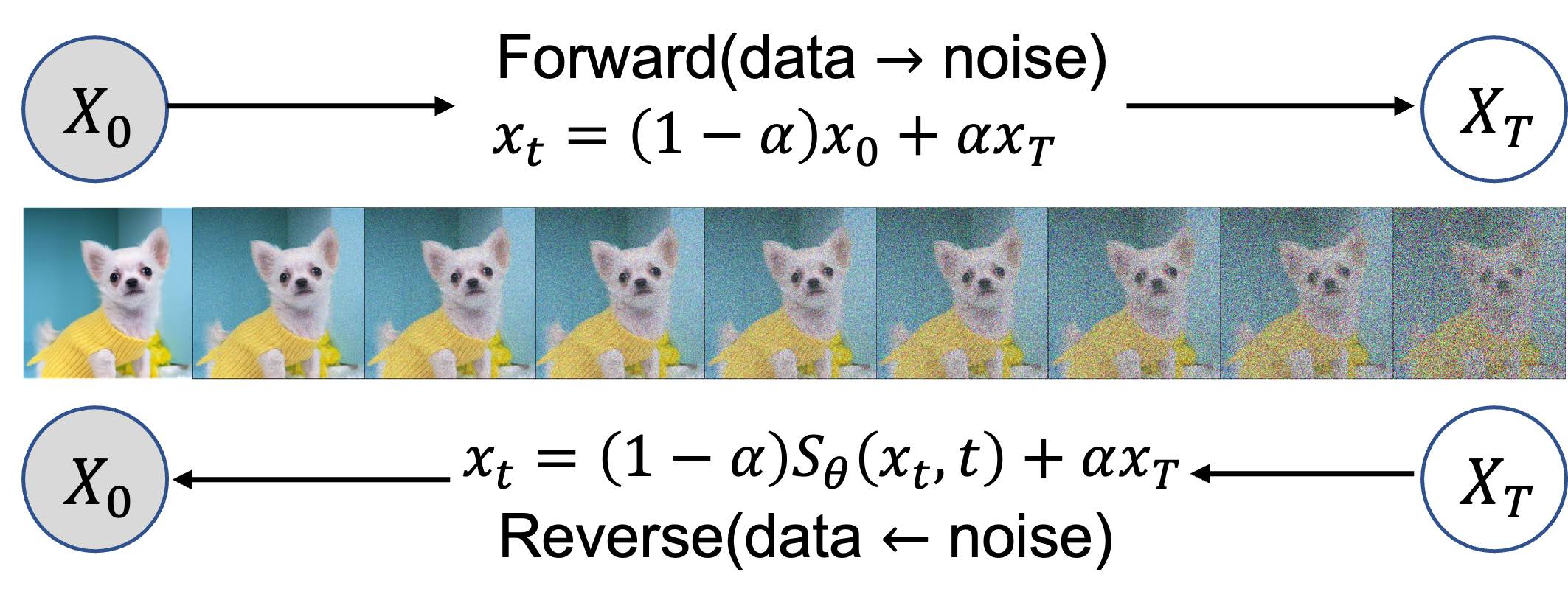}   
	\caption{Detail of the forward and reverse processes for the general diffusion model.     Standard diffusion add noise to the image until it is completely Gaussian                 noise, but the noise level cannot be controlled. We show that our general diffusion with linear interpolation model can control noisy added, finally match the real noise image, and the reverse process is step-by-step denoising.}   
	\label{fig1}    
\end{figure*}

Convolutional neural networks excel at extracting local information, but have limitations in capturing long-range dependencies between global data. The introduction of self-attention mechanisms~\cite{Vaswani2017AttentionIA}, residual feed-forward networks, and multi-head mechanisms in the Transformer architecture has revolutionized computer vision by effectively mining global interactions between textual information. Self-attention mechanisms~\cite{Cordonnier2019OnTR, Vaswani2021ScalingLS, Li2022DnSwinTR} can replace convolutional networks for convolutional-like operations and can represent more long-range correlations, and scaling techniques for local self-attention models have outperformed even efficient convolutional models. Image Transformer~\cite{Parmar2018ImageT} is a new image generator on the ImageNet dataset that performs well on ultra-high resolution tasks. ViT~\cite{Dosovitskiy2020AnII} is the representative work of transformer in the field of image recognition, which uses only the self-attention mechanism to achieve the current state-of-the-art recognition rate. Due to the excellent performance of transformer architecture on natural language and high-level vision tasks, Transformer has also been widely applied to real-world image denoising. Swin Transformer~\cite{Liu2021SwinTH} is a transformer-based backbone network that adapts to multi-scale images while reducing computational complexity. SwinIR~\cite{Liang2021SwinIRIR} consists of Transformer architecture, including three modules: shallow feature extraction, deep feature extraction, and high-quality image reconstruction, which can effectively remove the interference of severe noise while preserving high-frequency image details, resulting in sharper edges and more natural textures. SUNet~\cite{Ronneberger2015UNetCN} uses the Swin Transformer as a basic block and applies it to the UNet architecture for image denoising. Uformer~\cite{Wang2021UformerAG} designs the local enhancement window transformer module and the skip connection mechanism to show excellent performance in real-world image denoising. Eformer~\cite{Luthra2021EformerEE} first uses the transformer for medical image denoising and constructs an encoder-decoder network for medical image denoising through the transformer module. The transformer structure solves the deficiency of convolutional neural networks in extracting the global information feature of the image, and the self-attention mechanism can produce a more interpretable model. However, transformer modules are usually very large and computationally intensive, which cannot be applied to most image restoration tasks involving high-resolution images, and the acquisition of local information by the transformer is not as strong as convolutional neural networks in some cases.

Recent studies have shown that diffusion models~\cite{Ho2020DenoisingDP, Song2020DenoisingDI, SanRoman2021NoiseEF, Bansal2022ColdDI, Chen2022DiffusionDetDM} have achieved excellent performance in the field of generative modeling~\cite{Goodfellow2014GenerativeAN, Mirza2014ConditionalGA, Sohn2015LearningSO, Dinh2014NICENI}, even surpassing previous classical methods. However, although diffusion models have been widely used in other fields, they are still not widely used in the field of real-world image denoising. This is mainly because the final step image in the traditional diffusion model is Gaussian noise, and it is difficult to precisely control the level of noise added during the forward process.  While in the process of real-world image denoising, we hope that the last step image is the corresponding real-world noise image, and the level of added noise can be controlled, so as to gradually perform the denoising operation. Another reason is that real-world noise is extremely complex, which has various characteristics, including spatial and temporal dependence, frequency and color dependence, shot noise, pattern noise, and many other factors that contribute to the noise characteristics of different types of images. To address these problems, as shown in \figurename~\ref{fig1}, we propose a more general real-world image denoising diffusion model with linear interpolation, which achieves real-world image denoising through a forward process of gradual noise addition and a reverse process of gradual denoising operation. The intermediate noisy image is interpolated from the original clean image and the corresponding real-world noisy image, with the advantage that at time step $t=0$, the image is the original clean image, and at time step $t=T$, the image corresponds to the real-world noisy image. Our proposed method controls the amount of added noise by the parameter $\alpha=t/T$, so that the noise added in the diffusion process is closer to the real world than Gaussian noise with standard distribution, thus achieving more effective image denoising. Compared with traditional methods, this model performs well in various scenarios and has broad application prospects. In general, our contributions can be summarized as follows.

\begin{itemize}
\item Real-world image denoising using the diffusion model has not been studied, and we apply this method to real-world image denoising. It is noteworthy that our experimental results in image denoising show that using a simple CNNs Unet network alone can achieve denoising effects that approximate those of the Transformer architecture.
\item We propose a method using linear interpolation to precisely control the addition of noise to ensure that the added noise perfectly matches the noise image. With this method, the diffusion process can start with the original clean image and end with the corresponding real-world noise image instead of the pure Gaussian noise. This method provides new ideas for the field of image denoising and has the potential to achieve more precise and efficient image denoising operations in the future.
\item The effect of image denoising is strongly related to the sampling strategy adopted by the diffusion model. We propose an improved sampling algorithm with remarkable mathematical properties that can produce beneficial effects for real-world image denoising even when the original image cannot be accurately estimated. Compared with traditional sampling methods, our new method can achieve superior sampling results and bring new breakthroughs and innovations to the research and practice of image denoising.
\end{itemize}

\section{Background}

DDPM(Denoising Diffusion Probabilistic Model)~\cite{Ho2020DenoisingDP} is a type of probabilistic generative model based on diffusion processes. It models at the pixel level and can generate high-quality images. DDPM combines the affine transformation network and the diffusion process network to model the image distribution. The model uses the reverse diffusion algorithm as an optimization method and continuously improves its generative capability by iteratively optimizing the network parameters.

DDPM has a fixed Markov chain structure for the approximate posterior $q(\mathbf{x}_{1:T}|\mathbf{x}_0)$, which is also called the forward process or diffusion process. This Markov chain gradually adds Gaussian noise to the data over time, using a variance schedule $\beta_1, \ldots, \beta_T$ to control the amount of noise added at each point in the sequence:

\begin{equation}
q\left(\mathbf{x}_{1: T} \mid \mathbf{x}_0\right):=\prod_{t=1}^T q\left(\mathbf{x}_t \mid \mathbf{x}_{t-1}\right), \quad q\left(\mathbf{x}_t \mid \mathbf{x}_{t-1}\right):=\mathcal{N}\left(\mathbf{x}_t ; \sqrt{1-\beta_t} \mathbf{x}_{t-1}, \beta_t \mathbf{I}\right)
\end{equation}

The joint distribution $p_{\theta}(x_{0:T})$ is known as the reverse process in diffusion models, and it is defined as a Markov chain with learned Gaussian transitions. This chain starts with the prior distribution $p(x_T)= \mathcal{N} (x_T ; 0, I)$, which is also a Gaussian distribution with mean 0 and covariance matrix $I$. The Markov chain then reduces the noise according to the learned Gaussian transformation, gradually evolving over time until it reaches the observed data $x_0$:

\begin{equation}
p_\theta\left(\mathbf{x}_{0: T}\right):=p\left(\mathbf{x}_T\right) \prod_{t=1}^T p_\theta\left(\mathbf{x}_{t-1} \mid \mathbf{x}_t\right), \quad p_\theta\left(\mathbf{x}_{t-1} \mid \mathbf{x}_t\right):=\mathcal{N}\left(\mathbf{x}_{t-1} ; {\mu}_\theta\left(\mathbf{x}_t, t\right), \mathbf{\Sigma}_\theta\left(\mathbf{x}_t, t\right)\right)
\end{equation}

\section{General Diffusion Model}

The standard diffusion model that is commonly used in image generation tasks consists of two integral steps. First, a continuous random noise process is progressively added to the original image, resulting in a Gaussian distribution. A Unet model is then trained on this noisy image to estimate the added noise. Second, a pure Gaussian image is progressively denoised by subtracting the estimated noise from the image at each step using the same Unet model. This iterative process continues until the final version of the cleanly generated image is obtained. In our work, we present a more general diffusion model that takes into account matching the added noise to the real-world noisy image.

\begin{wrapfigure}{R}{0.495\textwidth}
    \begin{minipage}{0.495\textwidth}
        \begin{algorithm}[H]
            \caption{Training}\label{alg:training}
            \label{alg:sa1}
                \begin{algorithmic}[1]
                \Repeat
                  \State $x_0 \sim q(x_0), x_T \sim q(x_T)$
                  \State $t \sim \mathrm{Uniform}(\{0, \dotsc, T\})$
                  \State $\alpha=t / T$
                  \State $x_t=(1-\alpha) x_0+\alpha x_T$
                  \State Take gradient descent step on
                  \Statex $\qquad \nabla_\theta\left\|x_0-S_\theta\left(x_t, t\right)\right\|^2$
                \Until{converged}
                \end{algorithmic}
        \end{algorithm}
    \end{minipage}
\end{wrapfigure}

\subsection{Model Components and Training}
The study considers a clean image $x_0 \in \mathbb{R}^N$ and its corresponding real-world noisy image $x_T \in \mathbb{R}^N$, where a parameter $T$ determines the diffusion step. In Chapter 4 of the paper, the choice of $T$ is discussed as a hyperparameter that significantly influences the outcome of the process. To control the amount of noise added to the image, a parameter $\alpha$ is used, where $\alpha=t/T$. The Algorithm~\ref{alg:training} ensures that the noisy image is intermediate between $x_0$ and $x_T$, which differs from the conventional diffusion model that adds noise until a pure Gaussian image is obtained. The proposed method results in an improved overall quality of the real-world denoised image, while allowing control of the added noise. It is essential that the output distribution $x_t$ varies continuously with respect to $t$, and the operator must satisfy this requirement:

\begin{equation}
    x_t=(1-\alpha) x_0+\alpha x_T
\end{equation}

To implement a sequential reverse diffusion process, we use a basic Unet network called $S_\theta$. This network is specifically designed to accept two different inputs: the noisy image $x_t$ and the time step $t$, which allows the accurate estimation of the desired output $x_0$. The proposed architecture of the network facilitates the progressive denoising of the image by using deep learning driven reverse diffusion processing.

\begin{equation}
    x_t=(1-\alpha) S_\theta\left(x_t, t\right)+\alpha x_T
\end{equation}

In practical applications, a neural network parameterized by $\theta$ must be used. The objective function of this network is to accurately restore the image $x_0$, and this is achieved by minimizing the corresponding problem through training:

\begin{equation}
    \min_\theta \mathbb{E}_{x_0, x_T\sim p_{{data}}}\left\|S_\theta\left(x_t, t\right)-x_0\right\|
\end{equation}

where $x_0$ denotes a clean image randomly sampled from the distribution $p_{{data}}$,  $x_T$ denotes a real-world noisy image drawn from the same distribution. The intermediate image, denoted as $x_t$, is obtained by interpolating between $x_0$ and $x_T$. In the study, the norm used is represented by ‖·‖, which is specifically set to $\ell_2$ within the experimental framework.

Instead of using the $\mathcal{L}_1$ loss, our model is optimized with the robust Charbonnier loss~\cite{Charbonnier1994TwoDH} to better handle outliers and achieve better performance. The Charbonnier loss is defined as follows:

\begin{equation}
    \mathcal{L}_{\text\rm{char}}=\sqrt{\left\|I_{\text\rm{Denoised}}-I_{GT}\right\|^2+\varepsilon^2}
\end{equation}

where $I_{GT}$ represents the ground truth image and $\varepsilon$ is an empirical value, set to 0.001 in this paper. Compared to the $\mathcal{L}_1$ loss, the $\mathcal{L}_{\text\rm{char}}$ loss makes the model more robust.

\begin{wrapfigure}{R}{0.495\textwidth}
\begin{minipage}[t]{0.495\textwidth}
\begin{algorithm}[H]
  \caption{Origin Sampling} \label{alg:sampling1}
  \begin{algorithmic}[1]
    \vspace{.04in}
    \State $x_T \sim q(x_T)$
    \For{$t=T,T-1 \dotsc, 1$}
        \State $\quad \alpha=(t-1) / T$
        \State $\quad x_{t-1}=(1-\alpha) S_\theta\left(x_t, t\right)+\alpha x_T$
    \EndFor
    \State \textbf{return} $x_0$
    \vspace{.04in}
  \end{algorithmic}
\end{algorithm}
\end{minipage}
\begin{minipage}[t]{0.495\textwidth}
\begin{algorithm}[H]
  \caption{Improve Sampling} \label{alg:sampling2}
  \begin{algorithmic}[1]
    \vspace{.04in}
    \State $x_T \sim q(x_T)$
    \For{$t=T,T-1 \dotsc, 1$}
        \State $\quad \alpha_{t} = t / T$
        \State $\quad \alpha_{t-1} = (t-1) / T$
        \State $\quad\tilde{x}_t=\left(1-\alpha_t\right) S_\theta\left(x_t, t\right)+\alpha_t x_T$
        \State $\quad\tilde{x}_{t-1} =\left(1-\alpha_{t-1}\right) S_\theta\left(x_t, t\right)+\alpha_{t-1} x_T$
        \State $\quad x_{t-1}=x_t-\tilde{x}_t+\tilde{x}_{t-1}$
    \EndFor
    \State \textbf{return} $x_0$
    \vspace{.04in}
  \end{algorithmic}
\end{algorithm}
\end{minipage}
\end{wrapfigure}

\subsection{Sampling Algorithm}

The performance of real-world image denoising is highly dependent on the sampling strategy employed by the diffusion model, and the traditional method for generating an image using the diffusion model involves step-by-step sampling. This paper proposes a similar step-wise sampling algorithm for image denoising. In the forward process, a Unet is trained to estimate $x_0$ for images with different noisy levels obtained by interpolating between clean and real-world noisy images. In the reverse process, the real-world noisy image is denoised step-by-step using $S_{\theta}$. This corresponds to the standard sampling algorithm as described in Algorithm~\ref{alg:sampling1}.

The accuracy of Unet is better for real-world image denoising when $S_\theta\left(x_t, t\right)$ becomes equivalent to $x_0$. This is because as $S_\theta$ approaches $x_0$, the iterative results become more accurate for all steps of $t$. Therefore, the more accurate the estimation of $x_0$, the better the performance of real-world image denoising. However, the accuracy of Unet is generally not very good, and the errors begin to accumulate, leading to the deviation of the iteration results from $x_0$, resulting in poor image denoising performance.

The Unet is limited by the network architecture and data sizes, making it inaccurate in predicting $x_0$, which leads to error accumulation and causes imprecise iterative denoising effects (experimentally demonstrated in Section 4 of this paper). To address this problem, the Algorithm~\ref{alg:sampling2} is proposed to perform denoising sampling. The experimental results confirm that it outperforms the traditional sampling Algorithm~\ref{alg:sampling1} in terms of performance.

This sampling algorithm has remarkable mathematical properties that enhance the real-world image denoising effect beyond the Algorithm~\ref{alg:sampling1}. In particular, even if $S_\theta$ does not accurately estimate $x_0$, it can still produce a beneficial result for the real-world image denoising task. In the following section, we will elaborate on these special features and discuss their implications for improving real-world image denoising performance.

\subsection{Detail of Algorithm~\ref{alg:sampling2}}

It is clear from the two algorithms that if $S_\theta\left(x_t, t\right) = x_0 $ for all $t<T$, the two sampling algorithms can both denoise the real-world image accurately. In this section, we will analyze in detail the stability of these algorithms to errors in real-world image denoising.

The Algorithm \ref{alg:sampling1} can be written as the following formula\eqref{eq6}:

\begin{equation}\label{eq6}
\begin{aligned}
x_{t-1}&=(1-\alpha) S_\theta(x_t, t)+\alpha x_T\\
       &=[1-(t-1)/T]S_\theta(x_t, t)+(t-1)/T\cdot x_T=S_\theta(x_t, t)+(t-1)/T \cdot [x_T-S_\theta(x_t, t)]
\end{aligned}
\end{equation}

The Algorithm \ref{alg:sampling2} can be written as the following formula\eqref{eq7}:

\begin{equation}\label{eq7}
\begin{aligned}
x_{t-1}&=x_t-\tilde{x}_t+\tilde{x}_{t-1}\\
       &=x_t-(1-\alpha_t) S_\theta(x_t, t)-\alpha_t x_T+(1-\alpha_{t-1}) S_\theta(x_t, t)+\alpha_{t-1} x_T\\
       &=x_t-(1-t/T) \cdot S_\theta(x_t, t)-t/T \cdot x_T+[1-(t-1)/T] \cdot S_\theta(x_t, t)+(t-1)/T \cdot x_T\\
       &=x_t-1/T \cdot [x_T-S_\theta(x_t, t)]
\end{aligned}
\end{equation}

It can be seen from Equation~\ref{eq6} that the Algorithm~\ref{alg:sampling1} is more dependent on the result of $S_\theta(x_t, t)$, both the first and second terms of the formula are related to $S_\theta(x_t, t)$, and the coefficient of $x_T-S_\theta(x_t, t)$ is $t-1$ times larger than that of the Algorithm~\ref{alg:sampling2}, which will overestimate the error. If there is an error in the result of $S_\theta(x_t, t)$, after several rounds of iteration, the error will be infinite, resulting in poor performance for real world image denoising.

In contrast, the Algorithm~\ref{alg:sampling2} is more robust to erroneous predictions of $S_\theta(x_t, t)$. The first term has no direct relation to $S_\theta(x_t, t)$ and refers only to the current noise image $x_t$, the second term $x_T-S_\theta(x_t, t)$ is $t-1$ times smaller than that of Algorithm~\ref{alg:sampling1}. The positive or negative in front of the second term is not important, it is just the accumulation of errors under quantitative analysis.
From the Equation~\ref{eq6} and the Equation~\ref{eq7} we can conclude that the Algorithm~\ref{alg:sampling2} is less dependent on $S_\theta(x_t, t)$ and can handle larger errors more efficiently. After numerous iterations, the performance of the Algorithm~\ref{alg:sampling2} surpasses that of the Algorithm~\ref{alg:sampling1}, and we can draw the same conclusion from subsequent experiments.

\section{Experiments}

\begin{figure*}
	\centering	         
        \setlength{\abovecaptionskip}{-0.1cm}   
	\includegraphics[width=1\textwidth]{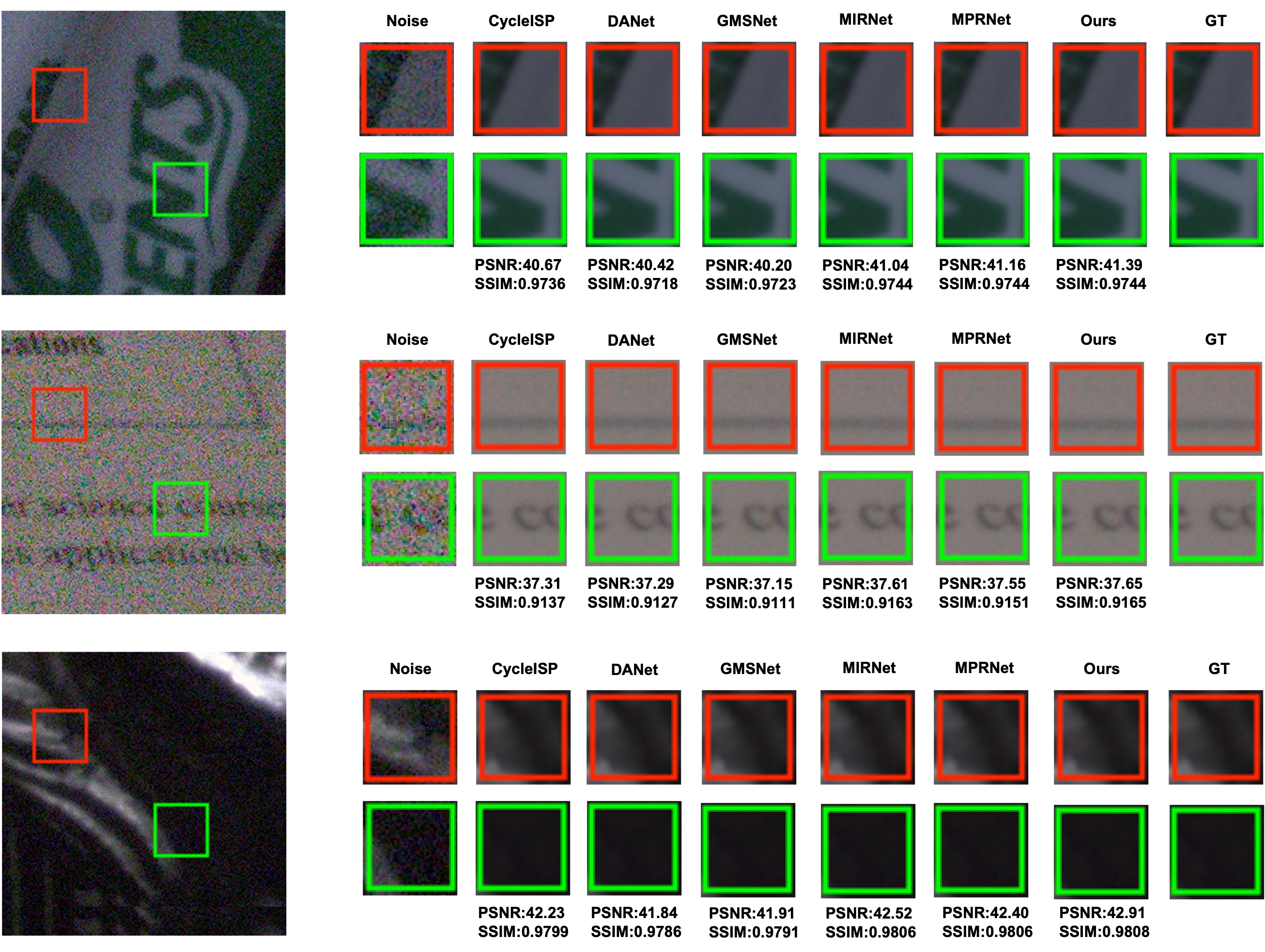}   
	\caption{Comparison with state-of-the-art methods on real noisy images from the SIDD validation dataset~\cite{Abdelhamed2018AHD}(\textbf{Zoom in for the best view}).}   
	\label{fig2}    
\end{figure*}

In this section, we will perform real-world image denoising experiments with our proposed method. First, we will introduce the real-world noisy image datasets and the evaluation index. Then, we will describe the experimental settings, including the training and testing settings and implementation details. Finally, we will compare our model with state-of-the-art methods on two real-world image denoising benchmarks.

\subsection{Datasets and Evaluation Index}

\begin{figure*}
	\centering	         
        \setlength{\abovecaptionskip}{0.1cm}   
	\includegraphics[width=1\textwidth]{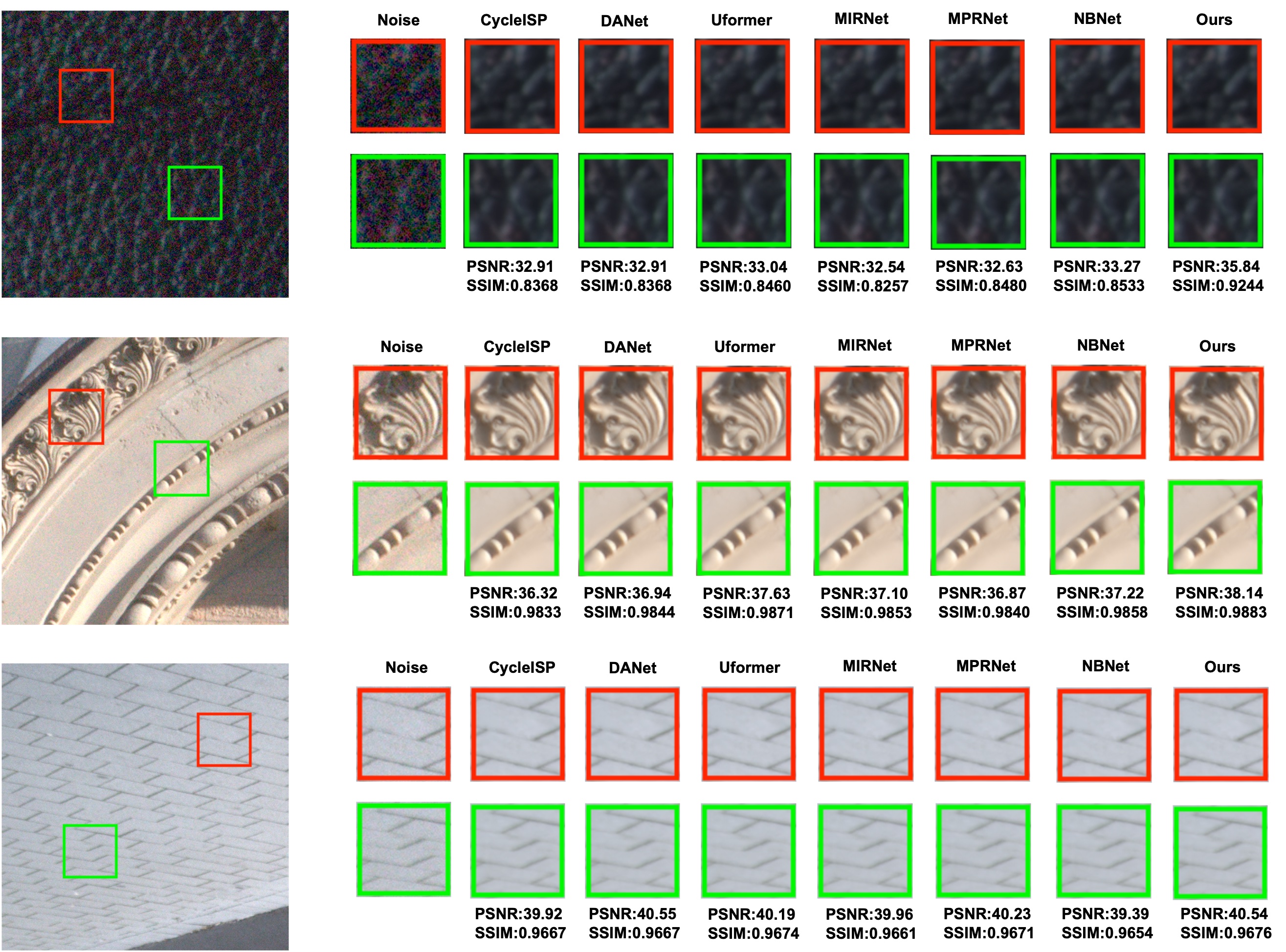}   
	\caption{Comparison with state-of-the-art methods on real noisy images from the DND benchmark~\cite{Pltz2017BenchmarkingDA}. Notice that no ground-truth images are provided by the DND benchmark(\textbf{Zoom in for the best view}).}   
	\label{fig3}    
\end{figure*}

To comprehensively verify our proposed method, we use the following two real-world noisy image datasets:

\paragraph{SIDD Dataset}The SIDD~\cite{Abdelhamed2018AHD} is a dataset for the evaluation of image denoising algorithms for smartphone cameras. It consists of 30,000 low-resolution (LR) images and their corresponding high-resolution (HR) counterparts, captured with five different smartphone cameras under different lighting conditions. The dataset comes with a predefined training and validation set, as well as an additional benchmark set for performance evaluation. For fast training and evaluation, we use the SIDD-Medium dataset (320 image pairs) for training and evaluate the method on the SIDD validation dataset\footnotemark[1].

\paragraph{DND Dataset}The DND~\cite{Pltz2017BenchmarkingDA} consists of 50 pairs of low-light scene images and corresponding high-quality reference images that have been captured under different imaging conditions. The dataset is intended to facilitate the evaluation of image denoising methods in real-world scenarios. Each image pair includes a raw noisy image, accompanied by its ground-truth clean version. The dataset's diverse range of scenarios makes it well-suited for challenging real-world denoising problems. However, the DND\footnotemark[2] does not provide any additional training data for fine-tuning denoising networks and can only be evaluated online.

\footnotetext[1]{https://www.eecs.yorku.ca/\~kamel/sidd/benchmark.php}
\footnotetext[2]{https://noise.visinf.tu-darmstadt.de/benchmark/\#results\_srgb}

In terms of evaluation metrics, we quantitatively analyze real-world image denoising using structural similarity (SSIM)~\cite{Wang2004ImageQA} and peak signal-to-noise ratio (PSNR)~\cite{Zhang2018TheUE}, which focus on pixel fields and are the most commonly used evaluation metrics in image restoration.

\subsection{Implementation Detail and Competing Methods}


Following the general training of GMSNet~\cite{Song2020GroupedMN}, we use the Adam optimizer~\cite{Kingma2014AdamAM} with $\beta_1=0.9$ and $\beta_2=0.999$ to train our model. To improve the trade-off between the size of the input patches and the available computing power, we set the batch size to 32 and the image patch size to $256\times256$. The learning rate is $2\times 10^{-4}$. For all experiments we use flipping and random rotation with angles of $90^{\circ}$, $180^{\circ}$ and $270^{\circ}$ for data augmentation. All experiments are performed in a Linux environment with PyTorch (1.12.0) running on a server with an NVIDIA RTX A6000 GPU. Nvidia CUDA 11.7 and cuDNN are used to accelerate the GPU computations. We use 500,000 iterations for the diffusion model, which takes about eight days to train.

To prove the superiority of our method, we compare it to state-of-the-art approaches on a real-world image denoising task. In particular, we evaluate its real-world image denoising performance on two test sets, the SIDD validation dataset~\cite{Abdelhamed2018AHD} and the DND benchmark~\cite{Pltz2017BenchmarkingDA}.

\paragraph{SIDD Validation Dataset} For the SIDD validation dataset, we follow the setting of DANet~\cite{Yue2020DualAN}, which uses only the SIDD-Medium dataset for training. We use three sampling algorithms for image denoising: Origin Sampling Algorithm~\ref{alg:sampling1}, Improved Sampling Algorithm~\ref{alg:sampling2}, and Direct Sampling Algorithm, which uses only one step $S_\theta\left(x_t, t\right)=x_0$ for image denoising. To evaluate the effectiveness of our proposed method, we compare it with a baseline model without the diffusion model, denoted as Without DDPM. We also compare our method with previously developed state-of-the-art methods, including CBDNet~\cite{Guo2018TowardCB}, VDN~\cite{Yue2019VariationalDN}, DANet~\cite{Yue2020DualAN}, MIRNet~\cite{Zamir2020LearningEF}, CycleISP~\cite{Zamir2020CycleISPRI}, MPRNet~\cite{Zamir2021MultiStagePI}, NBNet~\cite{Cheng2020NBNetNB}, GMSNet~\cite{Song2020GroupedMN}, Uformer~\cite{Fan2022SUNetST}, SwinIR~\cite{Liang2021SwinIRIR}.

\paragraph{DND} For the DND benchmark, we train our model on the SIDD Medium dataset and synthesize the noisy images provided by GMSNet~\cite{Song2020GroupedMN} for a fair comparison. The synthesized noisy images are generated by DIV2K~\cite{Agustsson2017NTIRE2C}. The high-resolution images in DIV2K are cropped into non-overlapping image patches, and then noise is added to these patches. We randomly sample 320 image patches from the SIDD-Medium dataset and 100 image patches from the synthetic noisy dataset for each training iteration. We also use three identical sampling algorithms, exactly the same as SIDD above and without DDPM, for image denoising, and use the same baselines as those implemented on the SIDD validation dataset for comparison purposes.

\begin{figure}
    \vspace{-0.1cm}
    \begin{minipage}{0.50\textwidth}
        \centering 
        \begin{tabular}{l|ccc}
            \toprule
            \multicolumn{1}{c|}{Methods}   & \multicolumn{1}{l}{PSNR$\uparrow$}   & \multicolumn{1}{l}{SSIM$\uparrow$}    \\ \midrule \midrule
            CBDNet~\cite{Guo2018TowardCB} & 30.78 & 0.801 \\
            VDN~\cite{Yue2019VariationalDN} & 39.28 & 0.956  \\
            DANet~\cite{Yue2020DualAN}        & 39.47 & 0.957  \\
            MIRNet~\cite{Zamir2020LearningEF} & 39.72   & 0.959  \\
            CycleISP~\cite{Zamir2020CycleISPRI}     & 39.52    & 0.957  \\
            MPRNet~\cite{Zamir2021MultiStagePI}  & 39.71  & 0.958  \\
            NBNet~\cite{Cheng2020NBNetNB}        & 39.75 & \underline{0.959} \\
            GMSNet~\cite{Song2020GroupedMN} & 39.63 & 0.956 \\ 
            Uformer~\cite{Fan2022SUNetST} &\underline{39.77} & \underline{0.959}  \\
            SwinIR~\cite{Liang2021SwinIRIR} &\underline{39.77} &0.958  \\\midrule \midrule
            Origin Sampling     & 34.93  & 0.922 \\
            Improve Sampling     & 39.40 & 0.957 \\
            Direct Sampling     & \textbf{39.81}  & \textbf{0.959} \\ 
            Without DDPM        &39.30    & 0.957  \\     \bottomrule
        \end{tabular}       
        \captionof{table}{\small{Quantitative results on the SIDD validation dataset~\cite{Abdelhamed2018AHD}.}}
        \label{tab:sidd}
    \end{minipage}\hfill
    \begin{minipage}{0.50\textwidth}
        \centering
        \begin{tabular}{l|ccc}
            \toprule
            \multicolumn{1}{c|}{Methods}   & \multicolumn{1}{l}{PSNR$\uparrow$}   & \multicolumn{1}{l}{SSIM$\uparrow$}   \\ \midrule \midrule
            CBDNet~\cite{Guo2018TowardCB} & 38.06 & 0.942 \\
            VDN~\cite{Yue2019VariationalDN} & 39.38 & 0.952  \\
            DANet~\cite{Yue2020DualAN}  & 39.58 & 0.955  \\
            MIRNet~\cite{Zamir2020LearningEF} & 39.88  & 0.956  \\
            CycleISP~\cite{Zamir2020CycleISPRI}     & 39.56    & 0.956    \\
            MPRNet~\cite{Zamir2021MultiStagePI}  & 39.80  & 0.954 \\
            NBNet~\cite{Cheng2020NBNetNB}        & 39.89 & 0.955  \\
            GMSNet~\cite{Song2020GroupedMN} & \underline{40.15} & \underline{0.961}  \\
            Uformer~\cite{Fan2022SUNetST} &39.96 & 0.956   \\ 
            SwinIR~\cite{Liang2021SwinIRIR} &40.01 &0.958  \\ \midrule \midrule
            Origin Sampling     & 38.14  & 0.943\\
            Improve Sampling     & 40.12  & 0.960 \\
            Direct Sampling     & \textbf{40.22}  & \textbf{0.962} \\ 
            Without DDPM        &40.08    & 0.960  \\ \bottomrule
        \end{tabular}
        \captionof{table}{\small{Quantitative results on the DND benchmark~\cite{Pltz2017BenchmarkingDA}.}}
        \label{tab:dnd}      
    \end{minipage}  
\end{figure}

\subsection{Comparison of Experimental Results}

\begin{table}[ht]
    \centering{
    \begin{tabular}{l|ccccccc}
    \toprule
    \multicolumn{1}{c|}{Time}  & 10 & 30 & 50 & 60 & 70 & 80 & 90         \\ \midrule \midrule
    \multicolumn{1}{c|}{PSNR} & 39.75 & 39.99 & 40.09 & 40.11 & \textbf{40.12} & 40.10 & 40.07 \\ \midrule
    \multicolumn{1}{c|}{SSIM} & 0.957 & 0.958 & 0.960 & 0.960 & \textbf{0.960} & 0.960 & 0.959 \\ \bottomrule
    \end{tabular}}
    \vspace{0.2cm}
    \caption{Quantitative results for time step from the DND benchmark dataset~\cite{Pltz2017BenchmarkingDA}.}
    \label{tab:time}
\end{table}

\paragraph{SIDD Validation Dataset} As depicted in Table~\ref{tab:sidd}, the results indicate that our model of Direct Sampling achieves better performance than all the state-of-the-art methods.  For example, our method compares favorably with the recently developed state-of-the-art SwinIR~\cite{Liang2021SwinIRIR} method, where the PSNR and SSIM of the proposed results are 0.04 dB and 0.001 higher than those of SwinIR, respectively, justifying the effectiveness of our model. Compared with DANet~\cite{Yue2020DualAN}, which uses a GAN~\cite{Mirza2014ConditionalGA}, our model still achieves a 0.34 dB improvement in the PSNR index. Additionally, we present the visual comparison in \figurename~\ref{fig2}. We can see that our model recovers sharper and clearer images than those of the other methods, and is therefore more faithful to the ground truth, which verifies the effectiveness of the diffusion model for image denoising. We can also see that the performance of Improve Sampling is better than the Origin Sampling and Without DDPM methods, which has demonstrated the efficiency of the sampling algorithm~\ref{alg:sampling2}. Although the performance of Improve Sampling is not better than Direct Sampling, this is not important and our experiments demonstrate the improvement of the diffusion model on the real-world image denoising performance.

\paragraph{DND benchmark} In Table~\ref{tab:dnd}, our model of Direct Sampling achieves a significant improvement over other state-of-the-art methods. Compared to MPRNet~\cite{Zamir2021MultiStagePI}, our model shows performance gains of 0.42 dB and 0.008 in terms of PSNR and SSIM, respectively, justifying the notion that real-world image denoising via diffusion model makes our model more robust in handling information with different frequencies. Compared to the ViT-based methods~\cite{Dosovitskiy2020AnII}, Uformer~\cite{Fan2022SUNetST} and SwinIR~\cite{Liang2021SwinIRIR}, our model outperforms them by a PSNR margin of up to 0.10 dB. The qualitative results shown in \figurename~\ref{fig3} verify that our model achieves a significant visual quality improvement over CycleISP~\cite{Zamir2020CycleISPRI} with realistic texture and clear structure. Our result also shows that Improve Sampling is better than Origin Sampling, and demonstrates the efficiency of the diffusion model, the result is the same as the SIDD validation dataset.

\paragraph{Time Steps} As shown in Table~\ref{tab:time}, diffusion steps have an effect on the denoising of real-world images. We perform detailed experiments on different steps on the DND benchmark, since the model must be retrained for different numbers of diffusion steps, our experiments only perform different training for the DND benchmark, and the sampling method is Improve Sampling. The performance of real-world image denoising is best when $t=70$.

\section{Conclusion}

This study presents an effective method for real-world denoising based on a general diffusion model with linear interpolation. The proposed method combines the advantages of the simple Unet and the diffusion model, which not only takes advantage of the local receptive field of CNNs for processing large images, but also exploits the generative advantage of the diffusion model. Specifically, in the forward process, the method interpolates images with different noise levels to estimate their noise using a simple Unet. In the reverse process, the model estimates and removes the noise step by step, resulting in real image denoising. We also propose two sampling algorithms and compare the results of the two sampling algorithms with Direct Sampling. Although the results of the sampling algorithms are not better than Direct Sampling, this limitation could be attributed to the insufficient rigidity of the diffusion model and is not a major concern. More importantly, the proposed method shows significant improvements in real-world image denoising performance.

Let's rethink, why does our method work? Through experiments, we have found that Direct Sampling is better than other sampling algorithms, we have also come to the conclusion that our proposed general diffusion model works in a manner similar to data augmentation. By interpolating noise images of different sizes, using the different time steps $(T)$ to control the degree of data augmentation, we can augment the dataset and use it to train a simple Unet. Data augmentation is widely used in machine learning as a technique to increase the size of training datasets, which can improve the generalization ability of models. Our method exploits this concept by using the diffusion process to generate new instances of noisy images with varying degrees of corruption. By using these augmented inputs during training, our model becomes more robust to variations in noise and can achieve better performance for real-world image denoising, which is also what makes Direct Sampling superior to other sampling algorithms.

Of course, our proposed method has some limitations, one of which is the imprecision of the diffusion process. This can affect the performance of the two sampling methods we use and result in lower quality samples compared to Direct Sampling. Although we have made improvements to the sampling Algorithm~\ref{alg:sampling2}, the results are still not as good as we would like. Addressing the imprecision of the diffusion process and optimizing the sampling algorithm will be a priority for future work on this topic. One possible approach to address this limitation would be to explore alternative diffusion models that provide greater precision and control over the samples generated. Another approach could be to improve the training process by incorporating additional loss functions or regularization techniques that encourage the model to generate higher quality samples.

The effectiveness of the method is demonstrated through extensive experiments, which include comparisons with state-of-the-art denoising methods on standard benchmark datasets. The results show that the proposed method achieves superior performance on various metrics, including PSNR and SSIM. Furthermore, the present work contributes to the field of image processing by advancing the understanding of how diffusion models can be integrated with CNNs to improve their performance in tasks such as denoising. The proposed method has potential applications in diverse domains, ranging from biomedical imaging to autonomous driving systems. We believe that our proposed method has the potential to be widely adopted in various domains due to its superior performance and versatility in handling different types of noise. Overall, we hope that our work will inspire further research in this area and lead to new innovations that can benefit society.

\section*{Acknowledgments}
This work is supported in part by the National Key R\&D Program of China (no. 2018AAA0100301), National Natural Science Foundation of China (no. 61976041), and Fundamental Research Funds for the Central Universities (DUT22LAB303).
 
\bibliographystyle{unsrt}  
\bibliography{main}  

\begin{thebibliography}{10}

\bibitem{Dong2014ImageSU}
Chao Dong, Chen~Change Loy, Kaiming He, and Xiaoou Tang.
\newblock Image super-resolution using deep convolutional networks.
\newblock {\em IEEE Transactions on Pattern Analysis and Machine Intelligence},
  38:295--307, 2014.

\bibitem{Schmidt2014ShrinkageFF}
Uwe Schmidt and Stefan Roth.
\newblock Shrinkage fields for effective image restoration.
\newblock {\em 2014 IEEE Conference on Computer Vision and Pattern
  Recognition}, pages 2774--2781, 2014.

\bibitem{Kim2015AccurateIS}
Jiwon Kim, Jung~Kwon Lee, and Kyoung~Mu Lee.
\newblock Accurate image super-resolution using very deep convolutional
  networks.
\newblock {\em 2016 IEEE Conference on Computer Vision and Pattern Recognition
  (CVPR)}, pages 1646--1654, 2015.

\bibitem{Nah2016DeepMC}
Seungjun Nah, Tae~Hyun Kim, and Kyoung~Mu Lee.
\newblock Deep multi-scale convolutional neural network for dynamic scene
  deblurring.
\newblock {\em 2017 IEEE Conference on Computer Vision and Pattern Recognition
  (CVPR)}, pages 257--265, 2016.

\bibitem{Jain2008NaturalID}
Viren Jain and H.~Sebastian Seung.
\newblock Natural image denoising with convolutional networks.
\newblock In {\em NIPS}, 2008.

\bibitem{Malfait1997WaveletbasedID}
Maurits Malfait and Dirk Roose.
\newblock Wavelet-based image denoising using a markov random field a priori
  model.
\newblock {\em IEEE transactions on image processing : a publication of the
  IEEE Signal Processing Society}, 6 4:549--65, 1997.

\bibitem{Burger2012ImageDC}
Harold~Christopher Burger, Christian~J. Schuler, and Stefan Harmeling.
\newblock Image denoising: Can plain neural networks compete with bm3d?
\newblock {\em 2012 IEEE Conference on Computer Vision and Pattern
  Recognition}, pages 2392--2399, 2012.

\bibitem{Vincent2008ExtractingAC}
Pascal Vincent, H.~Larochelle, Yoshua Bengio, and Pierre-Antoine Manzagol.
\newblock Extracting and composing robust features with denoising autoencoders.
\newblock In {\em International Conference on Machine Learning}, 2008.

\bibitem{Chen2015TrainableNR}
Yunjin Chen and Thomas Pock.
\newblock Trainable nonlinear reaction diffusion: A flexible framework for fast
  and effective image restoration.
\newblock {\em IEEE Transactions on Pattern Analysis and Machine Intelligence},
  39:1256--1272, 2015.

\bibitem{Mao2016ImageRU}
Xiao-Jiao Mao, Chunhua Shen, and Yubin Yang.
\newblock Image restoration using very deep convolutional encoder-decoder
  networks with symmetric skip connections.
\newblock In {\em NIPS}, 2016.

\bibitem{Cha2018FullyCP}
Sungmin Cha and Taesup Moon.
\newblock Fully convolutional pixel adaptive image denoiser.
\newblock {\em 2019 IEEE/CVF International Conference on Computer Vision
  (ICCV)}, pages 4159--4168, 2018.

\bibitem{Chen2018ImageBD}
Jingwen Chen, Jiawei Chen, Hongyang Chao, and Ming Yang.
\newblock Image blind denoising with generative adversarial network based noise
  modeling.
\newblock {\em 2018 IEEE/CVF Conference on Computer Vision and Pattern
  Recognition}, pages 3155--3164, 2018.

\bibitem{Goodfellow2014GenerativeAN}
Ian~J. Goodfellow, Jean Pouget-Abadie, Mehdi Mirza, Bing Xu, David
  Warde-Farley, Sherjil Ozair, Aaron~C. Courville, and Yoshua Bengio.
\newblock Generative adversarial nets.
\newblock In {\em NIPS}, 2014.

\bibitem{Zhang2016BeyondAG}
K.~Zhang, Wangmeng Zuo, Yunjin Chen, Deyu Meng, and Lei Zhang.
\newblock Beyond a gaussian denoiser: Residual learning of deep cnn for image
  denoising.
\newblock {\em IEEE Transactions on Image Processing}, 26:3142--3155, 2016.

\bibitem{Zhang2017FFDNetTA}
K.~Zhang, Wangmeng Zuo, and Lei Zhang.
\newblock Ffdnet: Toward a fast and flexible solution for cnn-based image
  denoising.
\newblock {\em IEEE Transactions on Image Processing}, 27:4608--4622, 2017.

\bibitem{Guo2018TowardCB}
Shi Guo, Zifei Yan, K.~Zhang, Wangmeng Zuo, and Lei Zhang.
\newblock Toward convolutional blind denoising of real photographs.
\newblock {\em 2019 IEEE/CVF Conference on Computer Vision and Pattern
  Recognition (CVPR)}, pages 1712--1722, 2018.

\bibitem{Vaswani2017AttentionIA}
Ashish Vaswani, Noam~M. Shazeer, Niki Parmar, Jakob Uszkoreit, Llion Jones,
  Aidan~N. Gomez, Lukasz Kaiser, and Illia Polosukhin.
\newblock Attention is all you need.
\newblock {\em ArXiv}, abs/1706.03762, 2017.

\bibitem{Cordonnier2019OnTR}
Jean-Baptiste Cordonnier, Andreas Loukas, and Martin Jaggi.
\newblock On the relationship between self-attention and convolutional layers.
\newblock {\em ArXiv}, abs/1911.03584, 2019.

\bibitem{Vaswani2021ScalingLS}
Ashish Vaswani, Prajit Ramachandran, A.~Srinivas, Niki Parmar, Blake~A.
  Hechtman, and Jonathon Shlens.
\newblock Scaling local self-attention for parameter efficient visual
  backbones.
\newblock {\em 2021 IEEE/CVF Conference on Computer Vision and Pattern
  Recognition (CVPR)}, pages 12889--12899, 2021.

\bibitem{Li2022DnSwinTR}
Hao Li, Zhijing Yang, Xiaobin Hong, Ziying Zhao, Junyang Chen, Yukai Shi, and
  Jin shan Pan.
\newblock Dnswin: Toward real-world denoising via continuous wavelet
  sliding-transformer.
\newblock {\em Knowl. Based Syst.}, 255:109815, 2022.

\bibitem{Parmar2018ImageT}
Niki Parmar, Ashish Vaswani, Jakob Uszkoreit, Lukasz Kaiser, Noam~M. Shazeer,
  Alexander Ku, and Dustin Tran.
\newblock Image transformer.
\newblock In {\em International Conference on Machine Learning}, 2018.

\bibitem{Dosovitskiy2020AnII}
Alexey Dosovitskiy, Lucas Beyer, Alexander Kolesnikov, Dirk Weissenborn,
  Xiaohua Zhai, Thomas Unterthiner, Mostafa Dehghani, Matthias Minderer, Georg
  Heigold, Sylvain Gelly, Jakob Uszkoreit, and Neil Houlsby.
\newblock An image is worth 16x16 words: Transformers for image recognition at
  scale.
\newblock {\em ArXiv}, abs/2010.11929, 2020.

\bibitem{Liu2021SwinTH}
Ze~Liu, Yutong Lin, Yue Cao, Han Hu, Yixuan Wei, Zheng Zhang, Stephen Lin, and
  Baining Guo.
\newblock Swin transformer: Hierarchical vision transformer using shifted
  windows.
\newblock {\em 2021 IEEE/CVF International Conference on Computer Vision
  (ICCV)}, pages 9992--10002, 2021.

\bibitem{Liang2021SwinIRIR}
Jingyun Liang, Jie Cao, Guolei Sun, K.~Zhang, Luc~Van Gool, and Radu Timofte.
\newblock Swinir: Image restoration using swin transformer.
\newblock {\em 2021 IEEE/CVF International Conference on Computer Vision
  Workshops (ICCVW)}, pages 1833--1844, 2021.

\bibitem{Ronneberger2015UNetCN}
Olaf Ronneberger, Philipp Fischer, and Thomas Brox.
\newblock U-net: Convolutional networks for biomedical image segmentation.
\newblock {\em ArXiv}, abs/1505.04597, 2015.

\bibitem{Wang2021UformerAG}
Zhendong Wang, Xiaodong Cun, Jianmin Bao, and Jianzhuang Liu.
\newblock Uformer: A general u-shaped transformer for image restoration.
\newblock {\em 2022 IEEE/CVF Conference on Computer Vision and Pattern
  Recognition (CVPR)}, pages 17662--17672, 2021.

\bibitem{Luthra2021EformerEE}
Achleshwar Luthra, Harsh Sulakhe, Tanish Mittal, Abhishek Iyer, and
  Santosh~Kumar Yadav.
\newblock Eformer: Edge enhancement based transformer for medical image
  denoising.
\newblock {\em ArXiv}, abs/2109.08044, 2021.

\bibitem{Ho2020DenoisingDP}
Jonathan Ho, Ajay Jain, and P.~Abbeel.
\newblock Denoising diffusion probabilistic models.
\newblock {\em ArXiv}, abs/2006.11239, 2020.

\bibitem{Song2020DenoisingDI}
Jiaming Song, Chenlin Meng, and Stefano Ermon.
\newblock Denoising diffusion implicit models.
\newblock {\em ArXiv}, abs/2010.02502, 2020.

\bibitem{SanRoman2021NoiseEF}
Robin~San Roman, Eliya Nachmani, and Lior Wolf.
\newblock Noise estimation for generative diffusion models.
\newblock {\em ArXiv}, abs/2104.02600, 2021.

\bibitem{Bansal2022ColdDI}
Arpit Bansal, Eitan Borgnia, Hong-Min Chu, Jie Li, Hamideh Kazemi, Furong
  Huang, Micah Goldblum, Jonas Geiping, and Tom Goldstein.
\newblock Cold diffusion: Inverting arbitrary image transforms without noise.
\newblock {\em ArXiv}, abs/2208.09392, 2022.

\bibitem{Chen2022DiffusionDetDM}
Shoufa Chen, Pei Sun, Yibing Song, and Ping Luo.
\newblock Diffusiondet: Diffusion model for object detection.
\newblock {\em ArXiv}, abs/2211.09788, 2022.

\bibitem{Mirza2014ConditionalGA}
Mehdi Mirza and Simon Osindero.
\newblock Conditional generative adversarial nets.
\newblock {\em ArXiv}, abs/1411.1784, 2014.

\bibitem{Sohn2015LearningSO}
Kihyuk Sohn, Honglak Lee, and Xinchen Yan.
\newblock Learning structured output representation using deep conditional
  generative models.
\newblock In {\em NIPS}, 2015.

\bibitem{Dinh2014NICENI}
Laurent Dinh, David Krueger, and Yoshua Bengio.
\newblock Nice: Non-linear independent components estimation.
\newblock {\em CoRR}, abs/1410.8516, 2014.

\bibitem{Charbonnier1994TwoDH}
Pierre Charbonnier, Laure Blanc-F{\'e}raud, Gilles Aubert, and Michel Barlaud.
\newblock Two deterministic half-quadratic regularization algorithms for
  computed imaging.
\newblock {\em Proceedings of 1st International Conference on Image
  Processing}, 2:168--172 vol.2, 1994.

\bibitem{Abdelhamed2018AHD}
A.~Abdelhamed, Stephen Lin, and M.~S. Brown.
\newblock A high-quality denoising dataset for smartphone cameras.
\newblock {\em 2018 IEEE/CVF Conference on Computer Vision and Pattern
  Recognition}, pages 1692--1700, 2018.

\bibitem{Pltz2017BenchmarkingDA}
Tobias Pl{\"o}tz and Stefan Roth.
\newblock Benchmarking denoising algorithms with real photographs.
\newblock {\em 2017 IEEE Conference on Computer Vision and Pattern Recognition
  (CVPR)}, pages 2750--2759, 2017.

\bibitem{Wang2004ImageQA}
Zhou Wang, Alan~Conrad Bovik, Hamid~R. Sheikh, and Eero~P. Simoncelli.
\newblock Image quality assessment: from error visibility to structural
  similarity.
\newblock {\em IEEE Transactions on Image Processing}, 13:600--612, 2004.

\bibitem{Zhang2018TheUE}
Richard Zhang, Phillip Isola, Alexei~A. Efros, Eli Shechtman, and Oliver Wang.
\newblock The unreasonable effectiveness of deep features as a perceptual
  metric.
\newblock {\em 2018 IEEE/CVF Conference on Computer Vision and Pattern
  Recognition}, pages 586--595, 2018.

\bibitem{Song2020GroupedMN}
Yuda Song, Yunfang Zhu, and Xin Du.
\newblock Grouped multi-scale network for real-world image denoising.
\newblock {\em IEEE Signal Processing Letters}, 27:2124--2128, 2020.

\bibitem{Kingma2014AdamAM}
Diederik~P. Kingma and Jimmy Ba.
\newblock Adam: A method for stochastic optimization.
\newblock {\em CoRR}, abs/1412.6980, 2014.

\bibitem{Yue2020DualAN}
Zongsheng Yue, Qian Zhao, Lei Zhang, and Deyu Meng.
\newblock Dual adversarial network: Toward real-world noise removal and noise
  generation.
\newblock {\em ArXiv}, abs/2007.05946, 2020.

\bibitem{Yue2019VariationalDN}
Zongsheng Yue, Hongwei Yong, Qian Zhao, Lei Zhang, and Deyu Meng.
\newblock Variational denoising network: Toward blind noise modeling and
  removal.
\newblock {\em ArXiv}, abs/1908.11314, 2019.

\bibitem{Zamir2020LearningEF}
Syed~Waqas Zamir, Aditya Arora, Salman~Hameed Khan, Munawar Hayat,
  Fahad~Shahbaz Khan, Ming-Hsuan Yang, and Ling Shao.
\newblock Learning enriched features for fast image restoration and
  enhancement.
\newblock {\em IEEE Transactions on Pattern Analysis and Machine Intelligence},
  45:1934--1948, 2020.

\bibitem{Zamir2020CycleISPRI}
Syed~Waqas Zamir, Aditya Arora, Salman~Hameed Khan, Munawar Hayat,
  Fahad~Shahbaz Khan, Ming-Hsuan Yang, and Ling Shao.
\newblock Cycleisp: Real image restoration via improved data synthesis.
\newblock {\em 2020 IEEE/CVF Conference on Computer Vision and Pattern
  Recognition (CVPR)}, pages 2693--2702, 2020.

\bibitem{Zamir2021MultiStagePI}
Syed~Waqas Zamir, Aditya Arora, Salman~Hameed Khan, Munawar Hayat,
  Fahad~Shahbaz Khan, Ming-Hsuan Yang, and Ling Shao.
\newblock Multi-stage progressive image restoration.
\newblock {\em 2021 IEEE/CVF Conference on Computer Vision and Pattern
  Recognition (CVPR)}, pages 14816--14826, 2021.

\bibitem{Cheng2020NBNetNB}
Shen Cheng, Yuzhi Wang, Haibin Huang, Donghao Liu, Haoqiang Fan, and Shuaicheng
  Liu.
\newblock Nbnet: Noise basis learning for image denoising with subspace
  projection.
\newblock {\em 2021 IEEE/CVF Conference on Computer Vision and Pattern
  Recognition (CVPR)}, pages 4894--4904, 2020.

\bibitem{Fan2022SUNetST}
Chi-Mao Fan, Tsung-Jung Liu, and Kuan-Hsien Liu.
\newblock Sunet: Swin transformer unet for image denoising.
\newblock {\em 2022 IEEE International Symposium on Circuits and Systems
  (ISCAS)}, pages 2333--2337, 2022.

\bibitem{Agustsson2017NTIRE2C}
Eirikur Agustsson and Radu Timofte.
\newblock Ntire 2017 challenge on single image super-resolution: Dataset and
  study.
\newblock {\em 2017 IEEE Conference on Computer Vision and Pattern Recognition
  Workshops (CVPRW)}, pages 1122--1131, 2017.

\end{thebibliography}


\end{document}